\documentclass[letterpaper, 10 pt, conference]{ieeeconf}  
\usepackage[utf8]{inputenc}
\usepackage[T1]{fontenc}

\IEEEoverridecommandlockouts                              

\usepackage{graphicx} 
\usepackage{float} 
\usepackage{amsmath} 
\usepackage{amssymb}  
\usepackage{algorithm} 
\usepackage{algpseudocodex}
\usepackage[font={small}]{caption}
\usepackage{subcaption}
\usepackage{booktabs}
\usepackage{makecell}
\usepackage{xspace}   

\let\citep\cite

\newcommand{\OURS}{\emph{\textbf Prior \textbf Reinforce}\xspace}
\newcommand{\ours}{\emph{\textbf P.\textbf R.}\xspace}

\title{\LARGE \bf
Prior Reinforce: Goal-Conditioned Dynamic Manipulation\\ with Limited Trials
}

\author{Yihang Hu$^{1,2*}$, Pingyue Sheng$^{1*}$, Yuyang Liu$^{1,2}$, Shengjie Wang$^{1,2}$, Yang Gao$^{1,2,3}$
\thanks{$^{*}$ Main Contributors.}%
\thanks{$^{1}$ IIIS, Tsinghua University.\;\;
$^{2}$ Shanghai Qi Zhi Institute.\;\;
$^{3}$ Spirit AI.}%
}

\begin{document}

\maketitle
\thispagestyle{empty}
\pagestyle{empty}




\begin{abstract}
Embodied robots have achieved strong performance in many real-world manipulation tasks, yet agile dynamic manipulation remains challenging due to high sensitivity to motion parameters and sparse outcome-level feedback. Tasks such as shooting a basketball into a hoop require precise control of fast open-loop motions, where small trajectory variations can lead to large outcome deviations, making data-efficient adaptation difficult for existing methods that rely on large-scale interaction, reward engineering, or accurate dynamic modeling. 
We propose \OURS (\ours), a simple and practical framework for goal-conditioned dynamic manipulation. The method first learns a structured motion manifold from a small set of demonstrations using a conditional diffusion model, and then adapts motions toward new goals through feedback-driven optimization in a low-dimensional condition space. By separating motion generation from outcome-driven adaptation, the framework enables efficient refinement using only a small number of real-world trials under noisy perception.
Experiments on multiple real-world dynamic manipulation tasks demonstrate that \ours reliably achieves new goals within as few as ten total trials while remaining robust to perception noise and hardware uncertainty, suggesting a practical approach for low-trial real-world robot adaptation.
\end{abstract}

\section{Introduction}
\label{sec:intro}

Robotic manipulation has advanced rapidly through reinforcement learning~\citep{luo2024serl, ye2023reinforcement}, imitation learning~\citep{bharadhwaj2023roboagent0, chi2023DP, chi2024umi, 
wang2025skil}, and large-scale vision-language-action models~\citep{black2024_0, kim2024openvla0, liu2024rdt1b, huang2024copa, lin2025onetwovla}. However, agile dynamic manipulation tasks remain challenging~\citep{chi2024IRP, huang2023dynamic}. Such tasks typically involve short-horizon open-loop execution, strong sensitivity to motion parameters, and sparse outcome-level feedback. Small variations in action trajectories can lead to large deviations in final results~\citep{zeng2020tossingbot}, making reliable adaptation under limited real-world trial budgets challenging.

Existing approaches often adapt or optimize motion directly in high-dimensional action or trajectory parameter spaces. Traditional analytical motion planning~\citep{sintov2015stochastic, taylor2019optimal} relies on accurate physical models, which are difficult to construct and brittle under uncertainty. Naive imitation learning~\citep{wang2025skil, chen2025tool, xu2024conditional}, such as behavior cloning, requires extensive demonstrations to cover the outcome space, while reinforcement learning~\citep{zeng2020tossingbot, munn2024wholebody} demands large interaction budgets and carefully designed reward functions. Although sim-to-real transfer~\citep{huang2023dynamic, munn2024wholebody} mitigates some costs, dynamic processes remain particularly sensitive to modeling errors. Residual refinement methods such as IRP~\citep{chi2024IRP} perform corrective updates in action space, which still requires searching over high-dimensional motion representations.

We propose an alternative perspective: instead of searching in action space, we perform adaptation in a learned low-dimensional motion manifold induced by task outcomes. For many dynamic manipulation problems, the dimensionality of goal-relevant results (e.g., 2D positional errors) is significantly lower than that of action trajectories. Within a coherent motion pattern, the mapping from a latent generation condition to the rollout result is often locally smooth. This structure suggests that adaptation in condition space can be substantially more sample-efficient than direct action-space refinement when only outcome-level feedback is available.

Based on this insight, we introduce \OURS (\ours), a two-stage framework. First, a conditional diffusion model learns a structured motion manifold from a small set of prior demonstrations. Second, given a new goal, a Gaussian process regression (GPR)–based adapter updates the generation condition using noisy semantic feedback from real-world rollouts. Instead of correcting raw action trajectories, \ours refines the generation condition within the learned motion manifold.
This separation between motion generation and semantic adaptation improves sample efficiency, enhances robustness to noisy feedback, and applies across multiple dynamic tasks without task-specific residual designs. We evaluate \ours on three real-world dynamic manipulation tasks and demonstrate that new goals can be reached within only 10 total trials. Simulator comparisons further indicate the effectiveness and efficiency of our two-stage approach compared to the baselines.
Our contributions are threefold:
\begin{itemize}
    \item We observe that many short-horizon dynamic manipulation tasks admit low-dimensional goal-relevant outcome spaces, motivating adaptation in condition space rather than direct action-space search.
    \item We propose a two-stage framework that learns a motion manifold via conditional diffusion and uses Gaussian process–based condition adaptation for goal refinement.
    \item We empirically show that condition-space adaptation enables effective goal reaching in both simulation and real-world settings under strict trial budgets.
\end{itemize}

\begin{figure*}[tb]
    \centering
    \vspace{10pt}
    \includegraphics[width=0.9\linewidth]{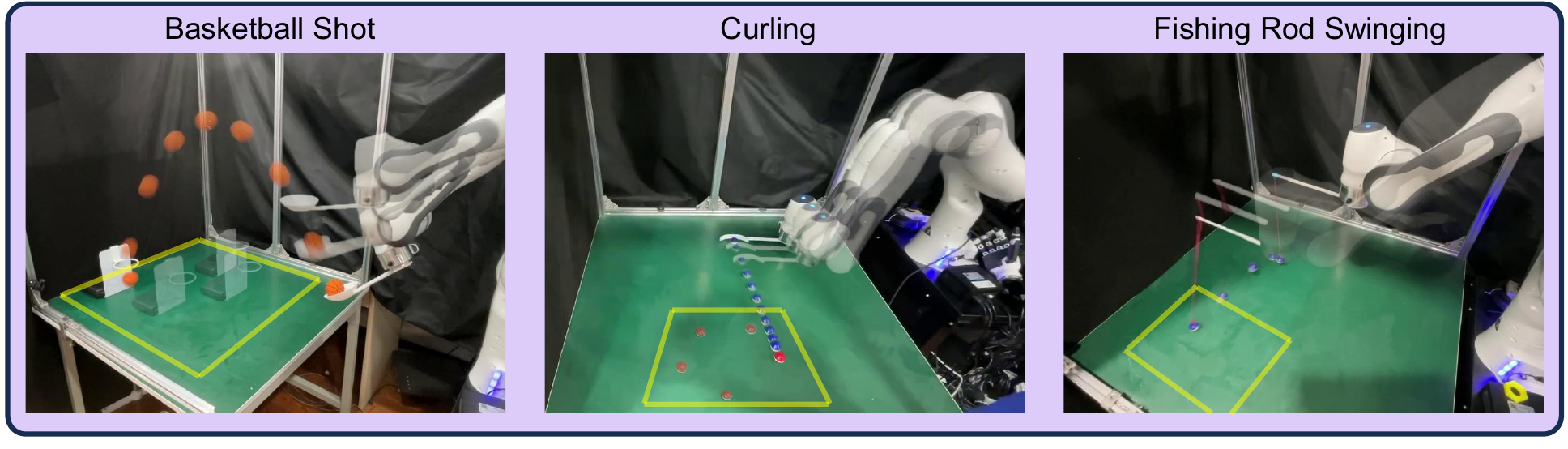}
    \vspace{-4pt}
    \caption{Overview of task layouts \& processes. The table is a $1\,\mathrm{m}\times1\,\mathrm{m}$ square. The region highlighted with yellow lines indicates the reachable goal space (e.g., where the hoop can be placed).}
    \label{fig:tasks}
    \vspace{-14pt}
\end{figure*}

\section{Related Work}
\label{sec:related}

\subsection{Dynamic Manipulation Learning}

Robotic interaction with dynamically evolving objects remains challenging due to strong sensitivity to motion parameters and complex physical dynamics. Classical analytical approaches rely on explicit modeling and trajectory optimization~\citep{sintov2015stochastic, taylor2019optimal}, which are difficult to scale to real-world settings. Learning-based methods have therefore been explored for dynamic manipulation tasks such as throwing, catching, and dynamic handover~\citep{zeng2020tossingbot, huang2023dynamic, chi2024IRP}. Recent work further investigates latent-space inference and replanning for dynamic manipulation~\citep{noh2025latent}, as well as iterative policy improvement in low-data regimes~\citep{van2025context}. While these approaches demonstrate impressive performance, they typically rely on substantial interaction data, structured modeling assumptions, or task-specific adaptation mechanisms, limiting their applicability under strict real-world trial budgets.

\subsection{Motion Representation and Generation}

Learning structured motion representations from demonstrations has been widely studied. Dynamic Movement Primitives (DMPs) and their probabilistic variants encode motion patterns as parameterized dynamical systems~\citep{Saveriano2023DMPsurvey, paraschos2013promp, li2023prodmps}. More recently, diffusion-based policies have shown strong capability in modeling multimodal robot trajectories directly from demonstrations~\citep{chi2023DP}. Diffusion has also been explored for planning and trajectory optimization~\citep{pan2024model}. These approaches primarily focus on learning expressive motion generators or policies, enabling trajectory synthesis and generalization across conditions. However, the problem of efficiently adapting generated motions toward new outcome goals using sparse outcome-level feedback remains largely orthogonal to these modeling efforts.

\subsection{Feedback-Driven Adaptation and Optimization}

Several works refine robot behavior using feedback from task outcomes. Residual learning and policy refinement methods update actions or policies directly based on observed errors~\citep{silver2018residual, johannink2019residual, alakuijala2021residual, zeng2020tossingbot, chi2024IRP}, while reinforcement learning approaches optimize policies through repeated interaction~\citep{munn2024wholebody}. These methods primarily operate in high-dimensional action or policy spaces, which can limit adaptation efficiency when only sparse outcome-level feedback is available.
In contrast, our approach leverages a structured low-dimensional condition space induced by a learned motion manifold. Instead of refining actions directly, we learn a Gaussian process regression (GPR)–based condition adapter that incrementally aligns desired outcomes with generation conditions using rollout feedback. This formulation enables feedback-driven adaptation without direct search in high-dimensional action trajectories, targeting efficient goal adaptation under limited real-world trials.

\section{Real-World Tasks}
\label{sec:tasks}

\OURS studies a class of goal-conditioned dynamic manipulation tasks frequently encountered in the real world. 
To provide an intuitive understanding before introducing the algorithm in Section~\ref{sec:method}, 
we first describe the three representative real-world tasks constructed in our laboratory 
(Figure~\ref{fig:tasks}).

\begin{enumerate}
    \item \textbf{Basketball Shot:} 
    The robot arm throws a toy basketball into a hoop placed on the table using a spoon-shaped end effector that allows natural release of the ball, resembling a human shooting motion.

    \item \textbf{Curling:} 
    The robot arm uses a bar-shaped end effector to push a curling stone (blue) from a fixed starting position. 
    The stone slides and decelerates naturally due to friction, and the objective is to stop it precisely in front of a target stone (red).

    \item \textbf{Fishing Rod Swinging:} 
    The robot arm equipped with a rod-like end effector swings a soft rope with a small metal “hook” attached at its end, attempting to cast it onto a target location on the table. 
    The rope introduces delayed and flexible-body dynamics, making precise positioning challenging.
\end{enumerate}

These tasks are carefully selected to cover several typical agile dynamic behaviors in daily activities, including projectile motion, sliding contact under friction, and flexible-body swinging. 
Despite their different physical mechanisms, they share the following common characteristics:

\begin{itemize}
    \item They involve short-horizon dynamic processes whose outcomes are highly sensitive to motion parameters.
    \item Successful executions exhibit smooth and continuous action patterns.
    \item Humans can accomplish these tasks with a small number ($\sim$10) of trial-and-error attempts, suggesting that efficient adaptation from limited experience is possible.
\end{itemize}

\section{Method}
\label{sec:method}

\begin{figure*}[tb]
    \centering
    \vspace{5pt}
    \includegraphics[width=1.0\linewidth]{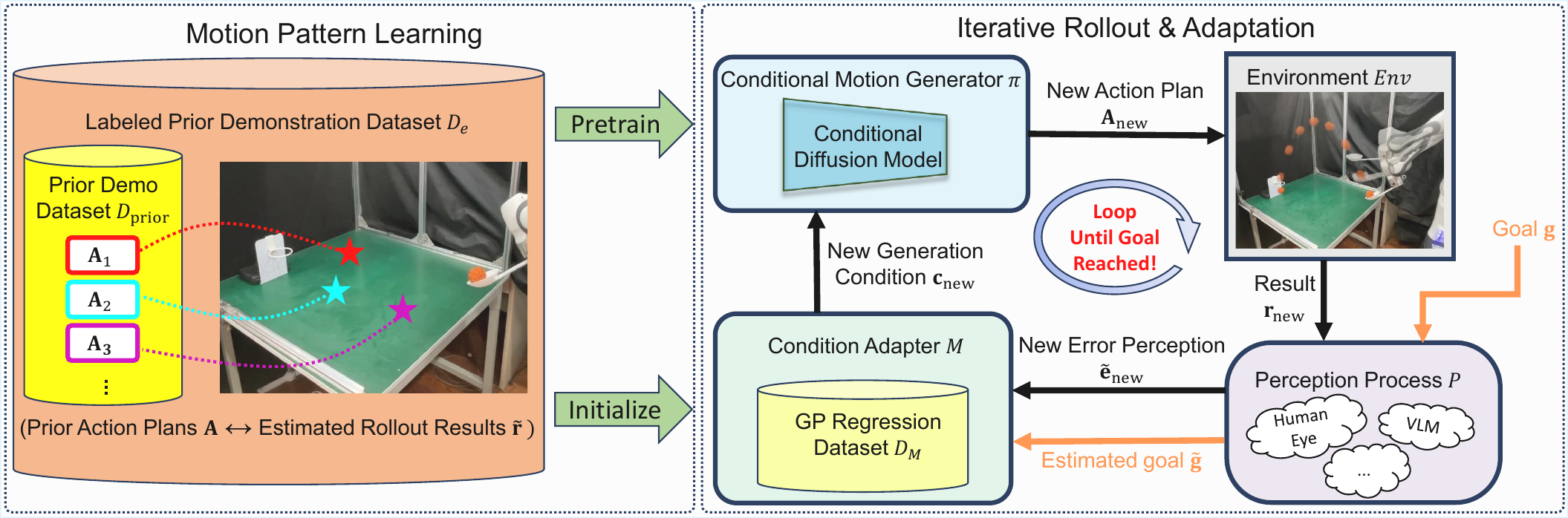}
    \vspace{-15pt}
    
    \caption{Overview of \OURS with its two stages: Motion Pattern Learning (left) and Iterative Rollout \& Adaptation (right).
    }
    \label{fig:adap}
    \vspace{-10pt}
\end{figure*}

\begin{algorithm}[t]
\caption{\OURS}
\label{algo:IDAP}
\begin{algorithmic}[1]
\Statex \textbf{Inputs:} Environment $Env$, perceived goal $\tilde{\mathbf g}$, perception model $P$, prior demonstrations $D_{\text{prior}}$, conditional motion generator $\pi$, condition adapter $M$.
\Statex

\Statex \textbf{Stage 1: Motion Pattern Learning}
\State Roll out demonstrations to construct dataset
\[
D_e=\{(\mathbf A_i,\tilde{\mathbf r}_i)\}.
\]
\State Train conditional diffusion generator $\pi:\tilde{\mathbf r}\rightarrow \mathbf A$ using $D_e$.

\Statex
\Statex \textbf{Stage 2: Iterative Rollout \& Adaptation}
\State Initialize adapter dataset
\[
D_M=\{(\mathbf c_i,\tilde{\mathbf r}_i)\mid \mathbf c_i=\tilde{\mathbf r}_i\}.
\]

\Repeat
    \State $\mathbf c^{new}=M(\tilde{\mathbf g})$
    \State $\mathbf A^{new}=\pi(\mathbf c^{new})$
    \State Execute rollout and observe $\tilde{\mathbf e}^{new}$
    \State Compute $\tilde{\mathbf r}^{new}=\tilde{\mathbf g}+\tilde{\mathbf e}^{new}$
    \State Update dataset $D_M \leftarrow D_M \cup \{(\mathbf c^{new},\tilde{\mathbf r}^{new})\}$
    \State Update adapter $M$ using $D_M$
\Until{$\|\tilde{\mathbf e}^{new}\|\le\epsilon\quad$ (Task succeeds.)}

\State \textbf{Output:} final action plan $\mathbf A_f$
\end{algorithmic}
\end{algorithm}

\subsection{Problem Formulation}
\label{subsec:formulation}

We formalize a class of goal-conditioned dynamic manipulation tasks characterized by short-horizon open-loop execution and outcome-level feedback. 
These tasks commonly arise in agile real-world interactions where the final outcome is determined primarily by the overall motion pattern rather than continuous feedback corrections during execution.

\subsubsection{Basic Assumptions}
\label{subsubsec:basic}

We consider dynamic manipulation tasks satisfying the following properties.

\textbf{Few Prior Motion Demonstrations:}
A small dataset of prior action plans is available, representing a coherent motion pattern shared across task instances. 
Such demonstrations define a restricted subspace of dynamically feasible motions, within which meaningful adaptation can be performed. 
This assumption reflects practical robotic settings where collecting large-scale interaction data for highly dynamic behaviors is costly, while a small number of demonstrations can already capture the essential motion structure.

\textbf{Semantic Feedback Perception:}
After each rollout, the agent receives a low-dimensional feedback vector that conveys task-relevant information with explicit physical meaning (e.g., positional error in task space). 
Instead of requiring full state observation or dense rewards, this feedback summarizes outcome quality and provides a direction for iterative refinement. 
Such outcome-level feedback is naturally obtainable in many real-world scenarios through simple measurements, visual estimation, or language-guided perception systems.

\textbf{Noisy Perception:}
The observed feedback obtained from vision-language models or human annotations is inevitably corrupted by perception noise, so the agent only has access to noisy estimates of outcome errors.
Moreover, dynamic manipulation involves inherent stochasticity in both physical execution and perception. Therefore, successful adaptation must remain robust under imperfect observations rather than relying on precise state estimation.

\textbf{Transient Open-loop Control:}
Dynamic manipulation is executed as short-horizon open-loop motion execution. 
The robot commits to a complete action sequence without intermediate feedback corrections during execution, which has limited influence on the final outcome due to fast dynamics, sensing inaccuracy and latency. 
Consequently, refinement across trials becomes more practical than closed-loop correction within a single rollout, motivating an outcome-driven adaptation formulation.

\subsubsection{Mathematical Formulation}

Let $Env$ denote a dynamic task environment, treated as an unknown black-box system that maps an action plan to a task outcome.
An action plan is defined as
\begin{equation}
\mathbf A=\{\mathbf a_t \mid t\in[H]\},
\end{equation}
where $\mathbf a_t$ denotes the robot action at timestep $t$ and $H$ is the horizon length.

Executing $\mathbf A$ in $Env$ produces a task outcome
\begin{equation}
\mathbf r = rollout_{Env}(\mathbf A), \qquad \mathbf r \in \mathbb R^k,
\end{equation}
where $\mathbf r$ represents a low-dimensional task-space result with explicit semantic meaning.

Given a desired goal $\mathbf g\in\mathbb R^k$, the true outcome error is
\begin{equation}
\mathbf e = \mathbf r-\mathbf g,
\end{equation}
while the agent observes a noisy perception
\begin{equation}
\tilde{\mathbf e}=P(\mathbf e),
\end{equation}
where $P$ denotes the perception process. Throughout the paper, variables with a tilde indicate noisy observations accessible to the agent.

The objective is to produce an action plan to achieve a new goal using minimal real-world trials:
\begin{equation}
\mathbf A_f = Algorithm(Env, D_{\text{prior}}, \mathbf g, P),
\end{equation}
where $D_{\text{prior}}=\{\mathbf A_i\}$ is a small set of prior demonstrations. The algorithm aims to produce an action plan whose perceived outcome error is sufficiently small within tolerance $\epsilon$ with respect to the specific task, for successful goal reaching:
\begin{equation}
\tilde{\mathbf e}=P\!\left(rollout(\mathbf A_f)-\mathbf g\right),
\qquad
\|\tilde{\mathbf e}\|\le \epsilon ,
\end{equation}

\subsection{Main Algorithm}
\label{subsec:main}

Based on the formulation above, our objective is to efficiently identify an action plan that reaches a newly specified goal using as few real-world trials as possible. 
Rather than directly optimizing high-dimensional action trajectories ($dim(\mathbf{A})=dim(\mathbf{a}_t)\cdot H$, about $7 * 140 \approx 10^3$ in our actual experiments), we leverage the assumptions stated above to design an adaptation framework consisting of two stages: \textbf{Motion Pattern Learning} and \textbf{Iterative Rollout \& Adaptation}. 
In the first stage, a structured motion manifold is learned from a small set of prior demonstrations. 
In the second stage, feedback-driven condition refinement is performed within this learned manifold until the desired goal is achieved. 
An overview of the pipeline is shown in Figure~\ref{fig:adap}, with pseudocode provided in Algorithm~\ref{algo:IDAP}.

\subsubsection{\textbf{Stage 1: Motion Pattern Learning}}
\label{stage1}

This stage learns a structured motion manifold from a small set of demonstrations. 
Each prior action plan $\mathbf A_i \in D_{\text{prior}}$ is executed once to obtain the perceived results of the rollout relative to a reference goal $\mathbf g_0$:
\begin{equation}
D_e=\{(\mathbf A_i,\tilde{\mathbf r}_i)\},
\quad
\tilde{\mathbf r}_i=\mathbf g_0 + P(rollout(\mathbf A_i)-\mathbf g_0).
\end{equation}

Using $D_e$, we train a conditional motion generator
\[
\pi:\tilde{\mathbf r}\rightarrow \mathbf A,
\]
which maps a desired outcome representation to an action plan.
This generator is implemented as a U-Net–based diffusion policy following~\citep{chi2023DP}. 
The condition vector is encoded using a two-layer MLP and injected into each U-Net block. 
Except for a slightly higher learning rate ($2\times10^{-4}$) to accelerate convergence on small datasets, all training settings follow the original formulation.

\subsubsection{\textbf{Stage 2: Iterative Rollout \& Adaptation}}
\label{stage2}

Given a new goal $\mathbf g$, directly using its perceived outcome $\tilde{\mathbf g}$ as the diffusion condition often leads to inaccurate outcomes due to sparse demonstrations and perception noise. 
Instead of directly modifying actions, \OURS adapts the generation condition through a learned \textit{Condition Adapter}
\[
M:\tilde{\mathbf r}\rightarrow \mathbf c,
\]
which models a mapping from observed rollout results to generation conditions. 
Intuitively, this learns an inverse relationship between outcomes and the conditions required to achieve them, compensating for systematic execution errors.

Let
$
\mathbf A_c=\pi(\mathbf c)
$
denote the generated action plan. 
The perceived outcome error is

\begin{equation}
\label{equ:egc}
\tilde{\mathbf e}(\mathbf c)
=
P(rollout(\pi(\mathbf c))-\mathbf g).
\end{equation}

During adaptation, each rollout produces a new observation
$(\mathbf c^{new},\tilde{\mathbf r}^{new})$ where
\begin{equation}
\tilde{\mathbf r}^{new}=\tilde{\mathbf g}+\tilde{\mathbf e}^{new}.
\end{equation}
These observations are accumulated into a dataset $D_M$, and the \textit{Condition Adapter} is re-trained online to update the mapping $M$.
Specifically, the \textit{Condition Adapter} is implemented as a Gaussian Process Regressor with a constant-scaled RBF kernel. (optimized with 50 restarts, and with a small diagonal regularization term for numerical stability.)

Besides, since real-world rollouts occasionally produce outliers due to hardware disturbances or perception failures, we additionally adopt a \textit{Data-Forgetting} strategy:
All Stage~1 samples are retained, while only the most recent $m$ adaptation trials are preserved in $D_M$ ($m=2$ in experiments). 
This prevents corrupted observations from permanently biasing the adapter while maintaining rapid adaptation.

\section{Experiments \& Evaluations}
\label{sec:experiments}

\subsection{Main Experiments in Real-world}
Our main evaluation focuses on real-world deployment, and this subsection introduces the task setups, success criteria, perception process, prior demonstration design, evaluation metric, and the corresponding experimental results.

\begin{figure}[b]
    \vspace{-15pt}
    \centering
    \includegraphics[width=1.01\linewidth]{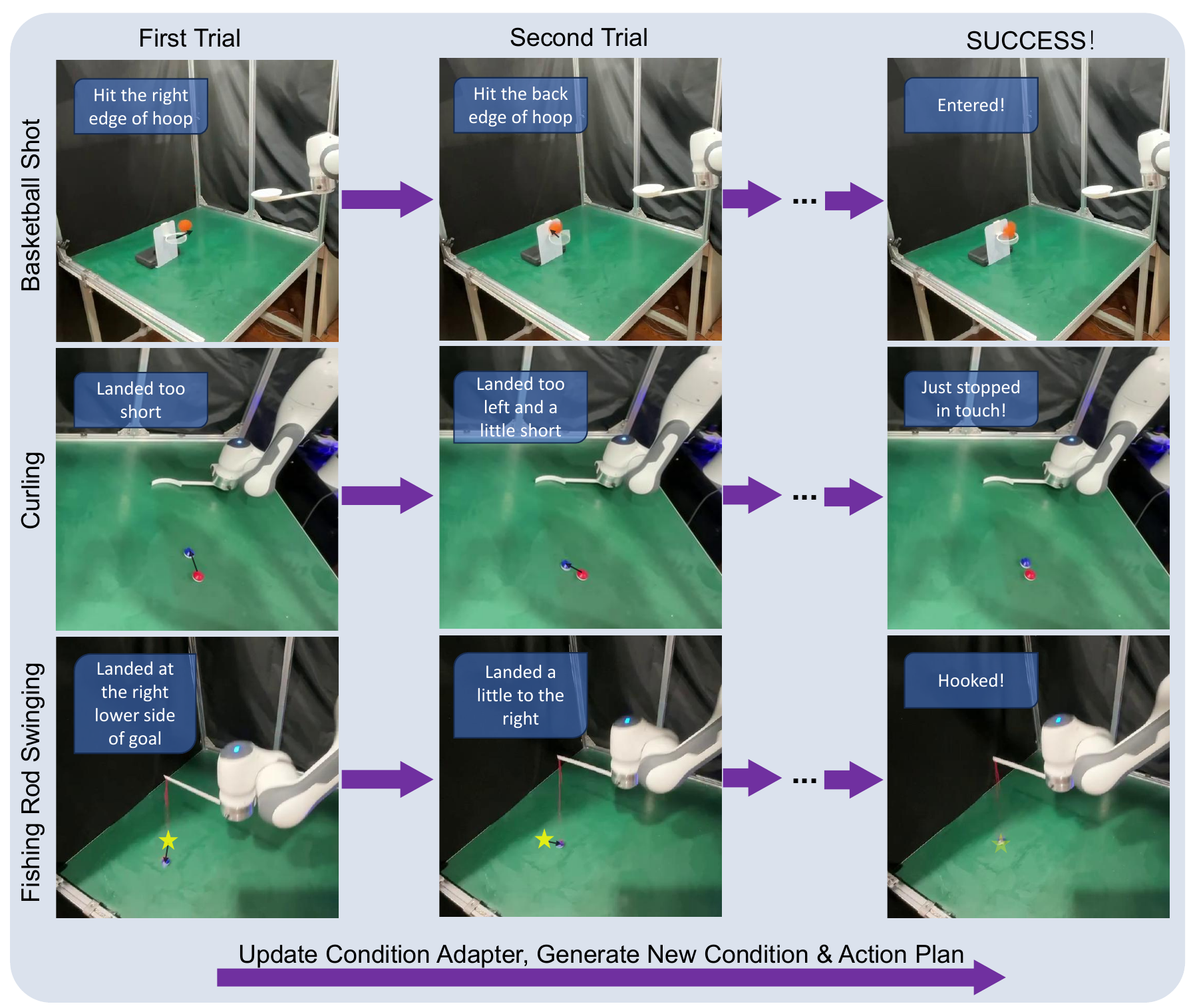}
    \vspace{-10pt}
    \caption{\textbf{Iterative Rollout \& Adaptation} process of \OURS on the three real-world tasks. 
    Each row shows one task. The small black arrows indicate the vectorized error estimated after an unsuccessful trial, which guides the next motion update. 
    The yellow star in \textit{Fishing Rod Swinging} denotes the goal location (a tiny magnet), which is visually small but requires precise contact.}
    \label{fig:exps}
\end{figure}

\begin{figure}[]
    \centering
    \vspace{5pt}
    \includegraphics[width=1\linewidth]{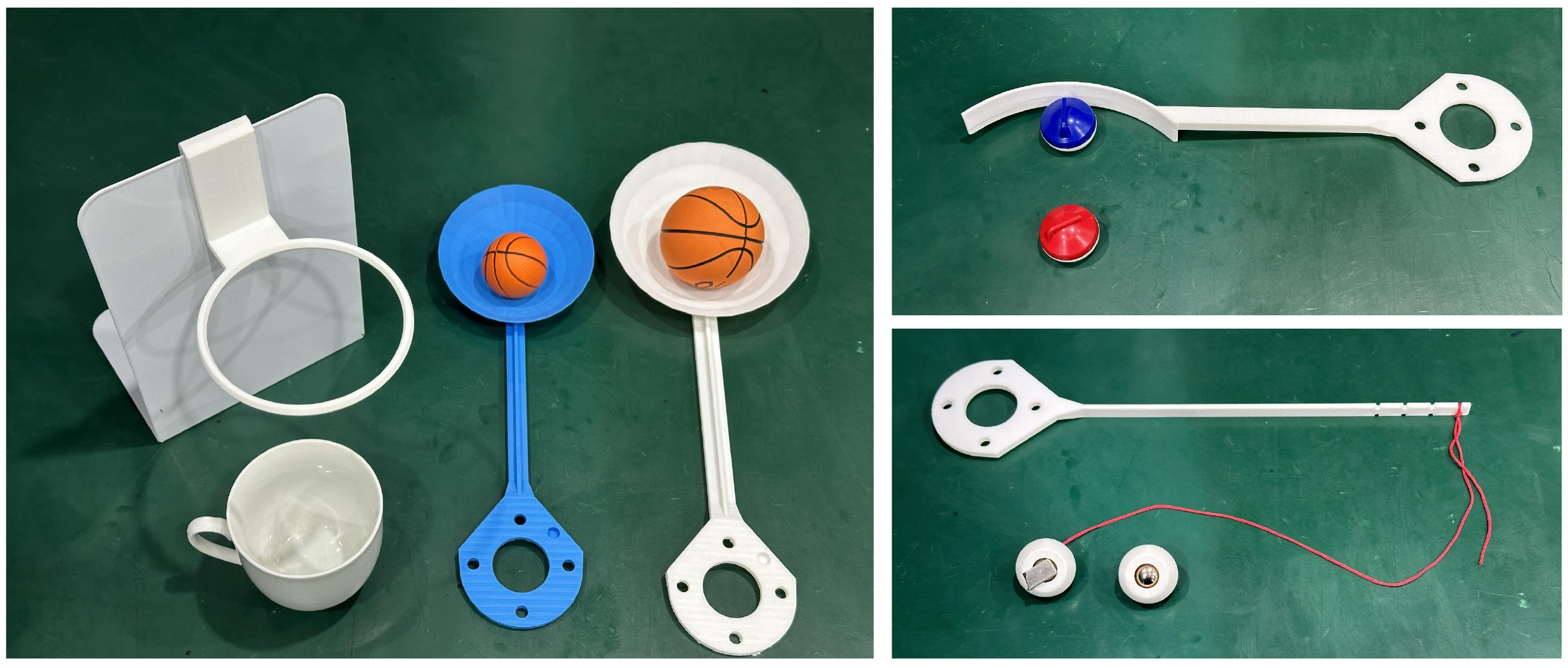}
    \caption{Hardware involved in real-world tasks:
    \textbf{Left:} Hardware for \textit{Basketball Shot} (\textit{V1}) and its 2 modified versions \textit{V2} and \textit{V3}. The original version \textit{V1} uses the hoop (diameter = 9.0 cm), white spoon to throw the bigger ball (diameter = 6.0 cm) into the hoop; \textit{V2} changes to the shorter blue spoon based on \textit{V1}; \textit{V3} changes to the smaller lighter ball and replace the hoop by a smaller cup, based on \textit{V2}.
    \textbf{Right Upper:} Hardware for \textit{Curling}.
    \textbf{Right Lower:} Hardware for \textit{Fishing Rod Swinging}. We borrow a curling stone from the last task to make use of the iron ball inside. The small magnet could only catch the iron ball in very close distance (about 1 cm).
    }
    \label{fig:hardware}
    \vspace{-14pt}
\end{figure}

\textbf{Task Setups:} 
We evaluate \OURS on the three real-world dynamic manipulation tasks introduced in Section~\ref{sec:tasks}, with the primary objective of measuring \emph{trial efficiency}, i.e., the number of real-world executions required to reach a previously unseen goal.
All tasks follow the assumptions in Section~\ref{subsubsec:basic} and are executed using a 7-dof Franka Emika Panda robot arm equipped with task-specific end effectors (Figure~\ref{fig:hardware}). 
Goals are arbitrarily placed on the 2D table surface within the reachable region of each task. 

\textbf{Success Criteria:} 
Strict precision criteria are adopted for all tasks.
For example, in \textit{Basketball Shot}, the hoop radius is only 1.5 times the ball radius, smaller in proportion than real basketball; in \textit{Fishing Rod Swinging}, the hook must collide with a tiny magnet within approximately 1\,cm to succeed.
These constraints lead to requirements for highly effective motion adaptation.

\begin{figure}[b]
    \vspace{-10pt}
    \centering
    \includegraphics[width=1.01\linewidth]{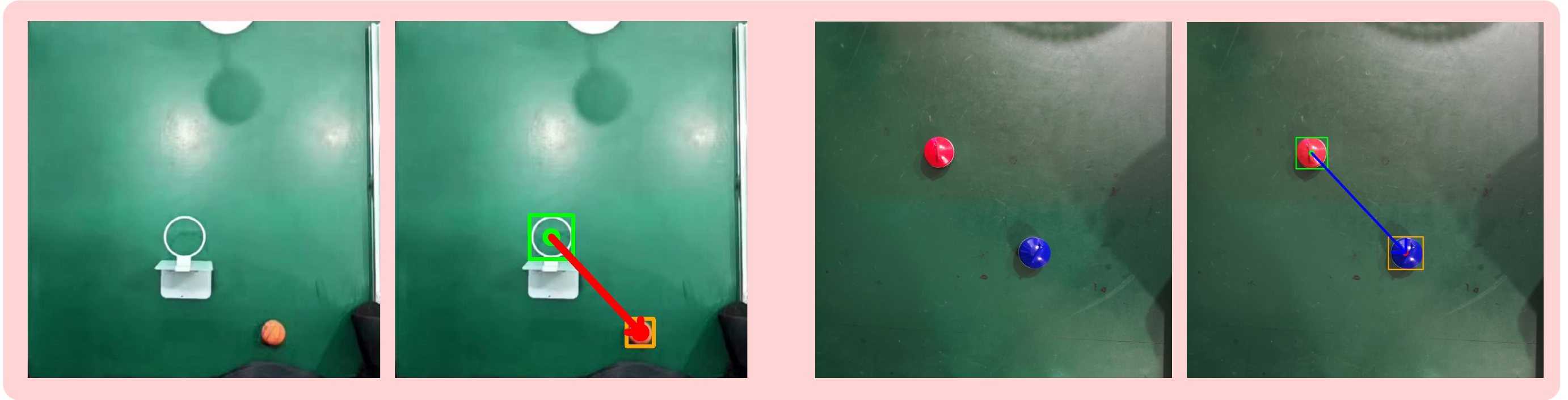}
    \vspace{-10pt}
    \caption{Example of object detection using GLM-4.5V. 
    The detected 2D positions of task-relevant objects are used to compute vectorized feedback for \textit{Basketball Shot} and \textit{Curling}.}
    \label{fig:VLM}
\end{figure}

\textbf{Perception Process:} 
Throughout our 3 tasks, the outcome $\mathbf{r}$ and condition $\mathbf{c}$ lie in the 2D space corresponding to the table surface plane.
For unsuccessful trials, feedback is defined as the vectorized positional difference between the reached point and the goal on the 2D table plane.
Perception is obtained either from human eye estimation or a prompted vision-language model (GLM-4.5V~\cite{vteam2025glm45v}), as illustrated in Figure~\ref{fig:VLM}.
We provide bird-eye-view images to the VLM and prompt it to detect bounding boxes of relevant objects in the key frame (collision frame or final stable frame).
Pixel differences between detected object centers are converted into feedback vectors. In terms of human feedback perception, the results are intentionally much coarser.
For example, in \textit{Basketball Shot}, distance estimates are discretized (0\,cm, 2\,cm, 5\,cm, multiples of 10\,cm).
Although VLM-based feedback is more precise, it may suffer from occasional outliers caused by irregular collisions or occasional recognition failures.
Importantly, \OURS achieves reliable convergence under both coarse human feedback and VLM perception.

\begin{table}[]
\vspace{5pt}
\caption{Settings of prior demonstration dataset $D_{\text{prior}}$ for each task. 
Newly generated action plans follow the same frame-action format as the priors.}
\vspace{-10pt}
\label{tab:prior}
\begin{center}
\begin{tabular}{@{}ccc@{}}
    \toprule
     \textit{Task}  & \textit{Prior Demo Source} &  \textit{Frame Action Format} \\
    \midrule
    \textit{Basketball Shot} & Simulator &  Joint pose \\
    \textit{Curling} & Tele-operation &  Joint pose \\
    \textit{Fishing Rod Swinging} & Tele-operation & End-effector pose \\
    \bottomrule
\end{tabular}
\end{center}
\vspace{-16pt}
\end{table}

\textbf{Prior Demonstrations:} 
Our framework does not constrain the source or format of demonstrations.
Task-specific configurations are summarized in Table~\ref{tab:prior}.
Depending on hardware limitations and task properties, demonstrations are collected either from simulation (for \textit{Basketball Shot}, using IsaacGym simulator) or via tele-operation (for \textit{Curling} and \textit{Fishing Rod Swinging}, using Meta Quest controller), depending on task-specific conditions. 
For example, tele-operation with our hardware cannot reliably generate sufficiently large acceleration for \textit{Basketball Shot}, motivating simulator-based prior data collection for that task.
Note that \ours itself does NOT depend on any simulator environment; the simulator environment of \textit{Basketball Shot} here is used only for collecting prior demonstrations. And for the other 2 tasks, no simulator environment is ever built throughout this paper.

The number of demonstrations is intentionally kept extremely small (6--8 per task) to reflect low-data conditions.
To ensure meaningful coverage, we distribute these demonstrations approximately across the reachable region shown in Figure~\ref{fig:tasks}, rather than clustering them around a single point.
All newly generated motions strictly follow the same frame-action format as the corresponding priors.

To further evaluate robustness, we construct two modified versions of \textit{Basketball Shot} (V2 and V3) with altered hardware configurations (Figure~\ref{fig:hardware}).
These variants directly reuse the same prior dataset collected for the original version, enabling cross-hardware generalization evaluation without additional data collection.

\textbf{Evaluation Metric:} 
The primary metric is the number of trials ($N$) required to first reach a new goal.
Stage~1 rollouts ($|D_{\text{prior}}|$) and average Stage~2 trials are summed to measure total interaction cost:
\[
N = |D_{\text{prior}}| + \text{average Stage~2 trials}.
\]
Success rate alone is not an appropriate metric in this setting, since repeating a successful motion typically yields high success probability regardless of adaptation efficiency.
Trial count directly measures how quickly the system converges to a valid solution for a new goal.

\begin{table}[]
\vspace{5pt}
\caption{Real-world experimental results. 
Fewer trials indicate better adaptation efficiency.}
\vspace{-10pt}
\label{tab:results}
\begin{center}
\begin{tabular}{@{}ccccc@{}}
    \toprule
     \textit{Task} & \textit{Perception} & \textit{Prior Num} & \textit{Trial Num} & \textit{Total} \\
     & & (Stage1) & (Stage2) & \\
    \midrule
    \textit{Basketball Shot} &eye & 6 & 2.9 $\pm$ 0.9 & 8.9 \\
    \textit{Basketball Shot} &VLM & 6 & 2.1 $\pm$ 1.1 & 8.1 \\
    \textit{Curling} &eye & 6 & 3.4 $\pm$ 1.1 & 9.4 \\
    \textit{Curling} &VLM & 6&  3.0 $\pm$ 1.2 & 9.0 \\
    \textit{Fishing Rod Swinging} &eye & 8 & 2.0 $\pm$ 0.8 & 10.0 \\
    \midrule
    \textit{Basketball Shot-V2} &eye & 6& 3.2 $\pm$ 0.7 & 9.2\\
    \textit{Basketball Shot-V3} &eye & 6& 3.7 $\pm$ 1.5 & 9.7\\
    \bottomrule
\end{tabular}
\end{center}
\vspace{-18pt}
\end{table}

\textbf{Real-world Experimental Results:}
For each task, two independent groups with different prior sets are evaluated, each containing five unseen goals.
Goals are deliberately placed far from prior rollout outcomes.

Across all tasks, a new goal is typically achieved within 2--4 iterative trials in Stage~2.
Including the 6--8 demo rollouts in Stage~1, as few as 10 total real-world trials are sufficient to reach the first unseen goal.
This reflects rapid trial-based adaptation like humans, where performance improves rapidly once a motion pattern is established.

VLM-based perception slightly reduces average trial counts in some tasks but introduces slightly higher variance due to perception outliers.
Despite hardware modifications in \textit{Basketball Shot V2} and \textit{V3}, performance degradation remains small, demonstrating that the learned motion manifold generalizes across similar task variants and that the iterative adaptation mechanism compensates for transfer gaps.

While applying \ours, convergence of training $\pi$ in Stage 1 takes about 10 min on an NVIDIA RTX 3090 GPU, and each re-training of $M$ in Stage 2 takes less than 0.5s, which is negligible during implementation.

\subsection{Comparisons \& Ablations in Simulator}

To systematically evaluate the effectiveness of each component in \OURS, 
we conduct controlled comparative experiments in simulation using the \textit{Basketball Shot} environment implemented in IsaacGym. 
The simulator enables large-scale parallel evaluation under identical initial conditions, 
allowing fair comparisons between our method and several representative ablative baselines.

We compare \OURS with the following methods, each sharing the same prior demonstration set $D_e$ (except for explicitly aligned variants):

\textbf{Baselines for Motion Generator:} The following baselines replace the \textit{Conditional Motion Generator} in \ours, while keeping the same GPR-based \textit{Condition Adapter} for fair comparison.
\begin{itemize}

    \item \textbf{Interpolation of Nearest Neighbors (I.N.N.).}
    I.N.N directly interpolates action plans whose rollout results are closest to the desired goal in $D_e$. 
    Interpolation weights are computed based on distances in result space. 

    \item \textbf{Dynamic Movement Primitives (DMP).}
    DMP models trajectories using boundary conditions and nonlinear forcing functions represented as weighted basis expansions. 
    We construct DMP representations from $D_e$ and interpolate the corresponding weight vectors to generate new action plans. 
    
    \item \textbf{UNet}. This baseline directly regresses from condition space to motion space using supervised neural network training (UNet) on the few prior samples.

\end{itemize}

\textbf{Baselines for Adapter:} The following baselines replace or modify the \textit{Condition Adapter} in \ours, while keeping the same diffusion-based \textit{Conditional Motion Generator} for fair comparison.
\begin{itemize}

    \item \textbf{\ours w/o Data Forgetting.}
    This ablative baseline retains all accumulated trials in $D_M$, i.e. $m=\infty$, evaluating the benefit of limiting historical data, especially under noisy perception. Note that we didn't choose other numbers for $m$ (say $m=4$) because it's hardly meaningful under our low trial budget.

    \item \textbf{\ours with Proportional Compensator.}
    Instead of GPR, 
    this variant updates the generation condition using a simple proportional correction rule:
    the next condition is obtained by subtracting $0.8$ times the last observed error from the last condition.
    
    \item \textbf{\ours w/o Adapter.} This variant freezes the generation condition, simply attempting to reach the goal by repeatedly generating actions and rolling them out.
\end{itemize}

\textbf{Additional Experimental Conditions.}
We also evaluate some of the above methods under two modified settings to enhance our experiments and analysis:

\begin{figure}[b]
    \vspace{-10pt}
    \centering
    \includegraphics[width=0.6\linewidth]{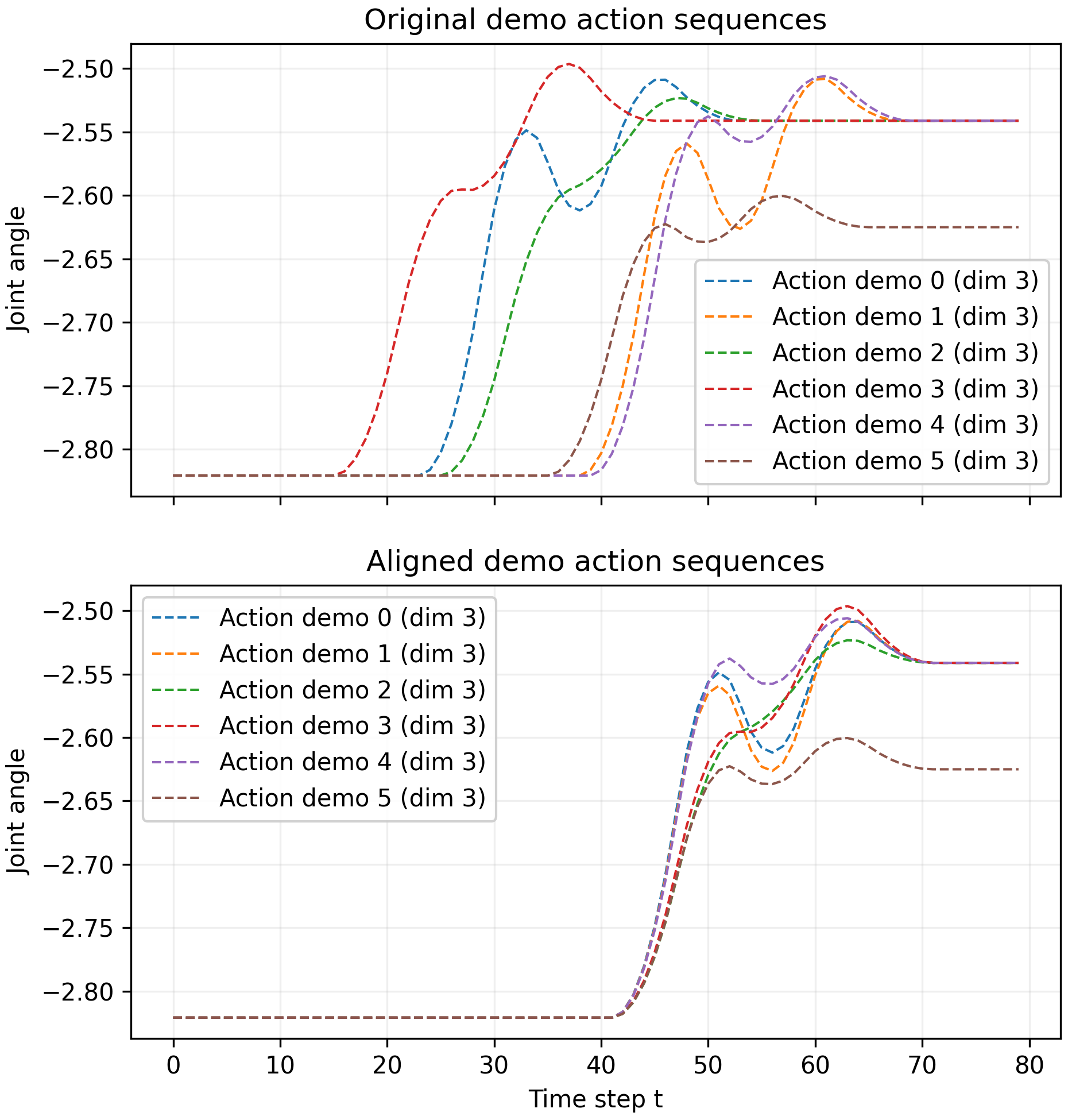}
    \vspace{-2pt}
    \caption{\textbf{Timeline Alignment} example: We plot original (top) and aligned (bottom) action demos (joint 3 pose) for \textit{Basketball Shot}.}
    \label{fig:align}
\end{figure}
\begin{itemize}

    \item \textbf{Timeline Aligned Variants.}
    In arbitrarily sourced prior demonstrations, the effective dynamic phase may begin at different timesteps across trajectories. 
    We additionally align the demonstrations so that the key dynamic phase occurs at similar timesteps (See Figure~\ref{fig:align}). 

    \item \textbf{Amplified Perception Noise.}
    Since simulator feedback is inherently more accurate than real-world perception, 
    we artificially amplify feedback noise by scaling each feedback vector with a random factor (0.7 to 1.4) and adding Gaussian noise. 
    This setting aims to approximate the uncertainty of real-world perceptions.

\end{itemize}

Note that we focus our comparisons on baselines that are compatible with the same low-demo and low-trial protocol, i.e., methods that can be instantiated from only a handful of prior demonstrations and adapted online using the same sparse outcome-level feedback. Accordingly, methods that require large-scale pretraining, many meta-training tasks, or richer supervision are not included as primary baselines.

\textbf{Evaluation Protocol.}
We uniformly sample 100 fixed goal locations within the reachable workspace and conduct parallel independent experiments for each goal. 
Performance is measured by success fractions across iteration rounds. 
(e.g., a value of 0.8 at round 2 indicates that 80\% of goals are reached in at most two trials, while 20\% are not.)
This metric reflects adaptation efficiency under limited trial budgets.

\begin{figure}[]
\begin{subfigure}{\linewidth}
    \centering
    \includegraphics[width=0.75\linewidth]{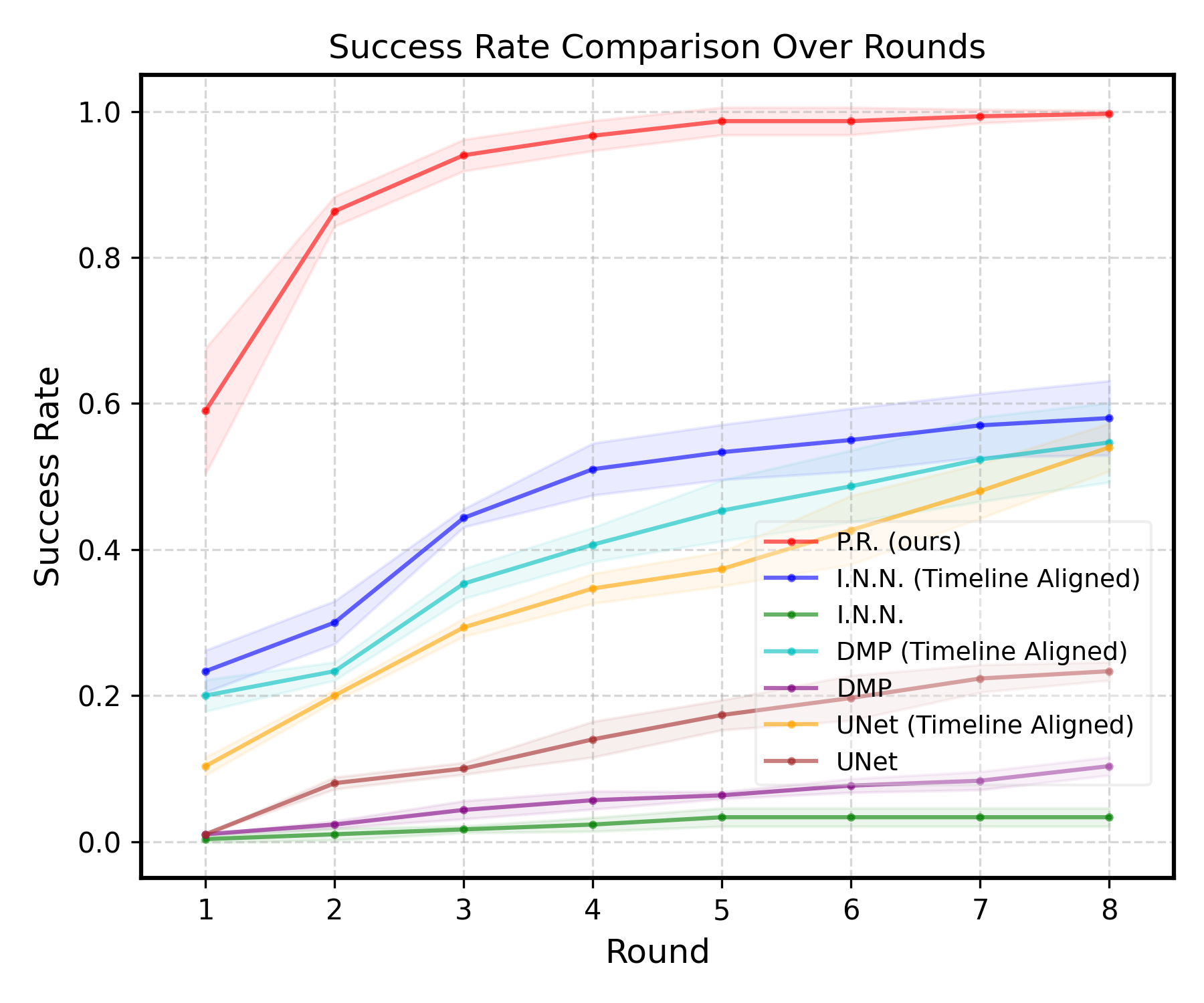}
    \vspace{-10pt}
    \caption{Success fractions of \ours and ablative baselines with replaced \textit{Conditional Motion Generator}.
    }
    \label{fig:compare}
\end{subfigure}


\begin{subfigure}{\linewidth}
    \centering
    \includegraphics[width=0.75\linewidth]{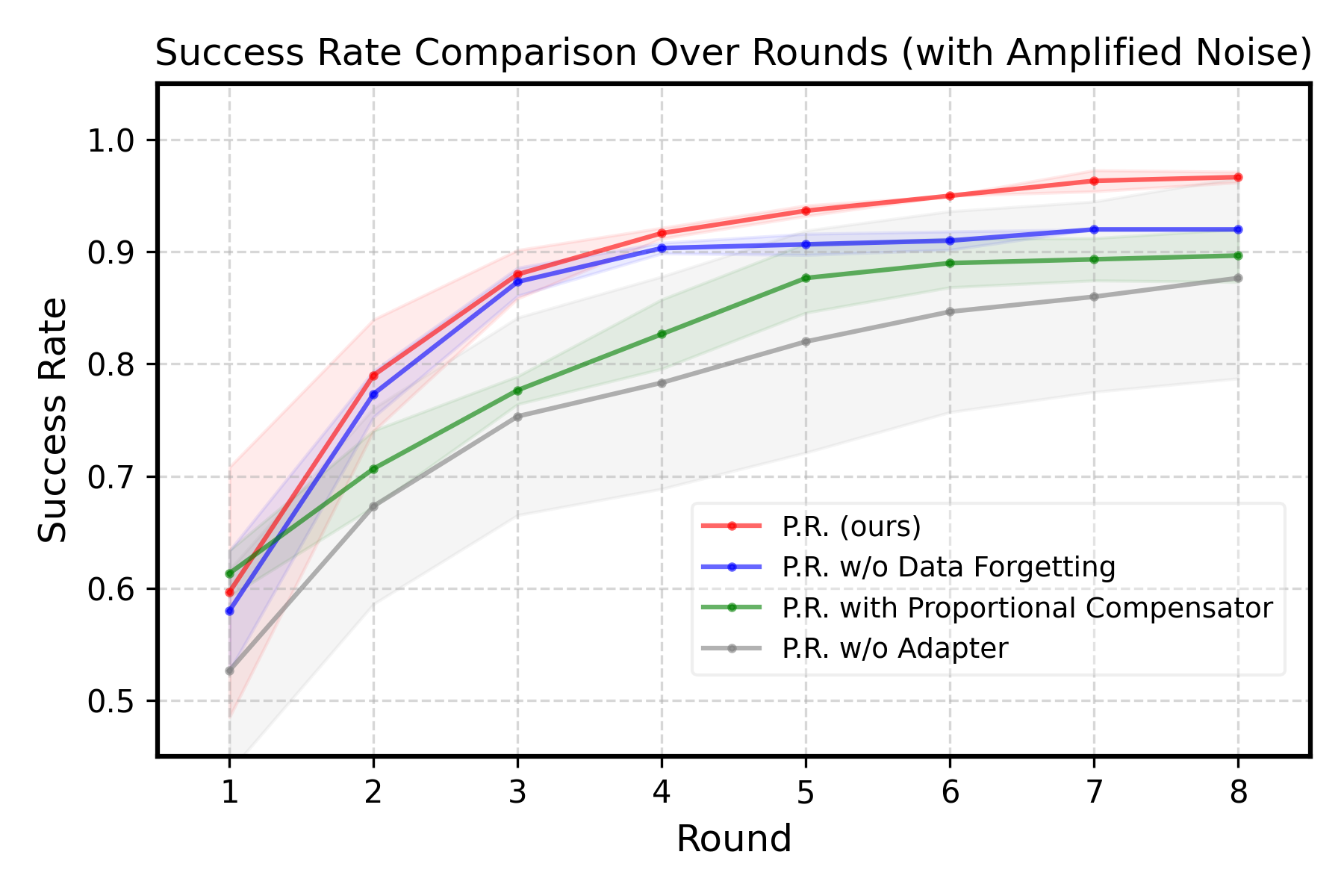}
    \vspace{-10pt}
    \caption{Success fractions of \ours and related ablative baselines with changed \textit{Condition Adapter}.
    All tested with the same level of \textbf{Amplified Perception Noise}, to mimic the real-world scenario.
    }
    \label{fig:compare_noise}
\end{subfigure}
\caption{Comparison results of \ours and ablative baselines.}
\label{fig:sim_results}
\vspace{-10pt}
\end{figure}

\textbf{Results and Analysis.}
Figure~\ref{fig:compare} compares \ours with baselines using alternative motion generators. 
All other three alternative generators struggle to capture the underlying motion structure when demonstrations are sparse and temporally misaligned. 
Manual timeline alignment significantly improves their performance, 
yet they remain consistently inferior to \ours, indicating that the diffusion-structure better captures coherent dynamic motion patterns beyond interpolation, classical parameterization, or regression-learning. 

Figure~\ref{fig:compare_noise} evaluates changed adapters under amplified perception noise. \ours with the GPR-based adapter reaches more than 90\% of goals within 4 rounds, whereas the Proportional Compensator exhibits slower convergence and lower asymptotic success rates (only marginally outperforms the no-adapter version). In addition, the Data-Forgetting strategy improves the upper performance bound, 
highlighting the importance of limiting historical data to mitigate the influence of outliers.

Overall, \OURS achieves over 90\% success within four additional trials even under challenging conditions 
where demonstrations are temporally misaligned and perception noise is amplified. 
These results confirm that both the learned motion manifold and the Gaussian-process-based condition adaptation are critical for efficient and robust goal-directed dynamic manipulation.

\subsection{Extra Experiments \& Visualizations}
\label{sec:extra_viz}

\begin{figure}[t]
    \centering
    \begin{subfigure}{0.7\linewidth}
        \centering
       \includegraphics[width=\linewidth]{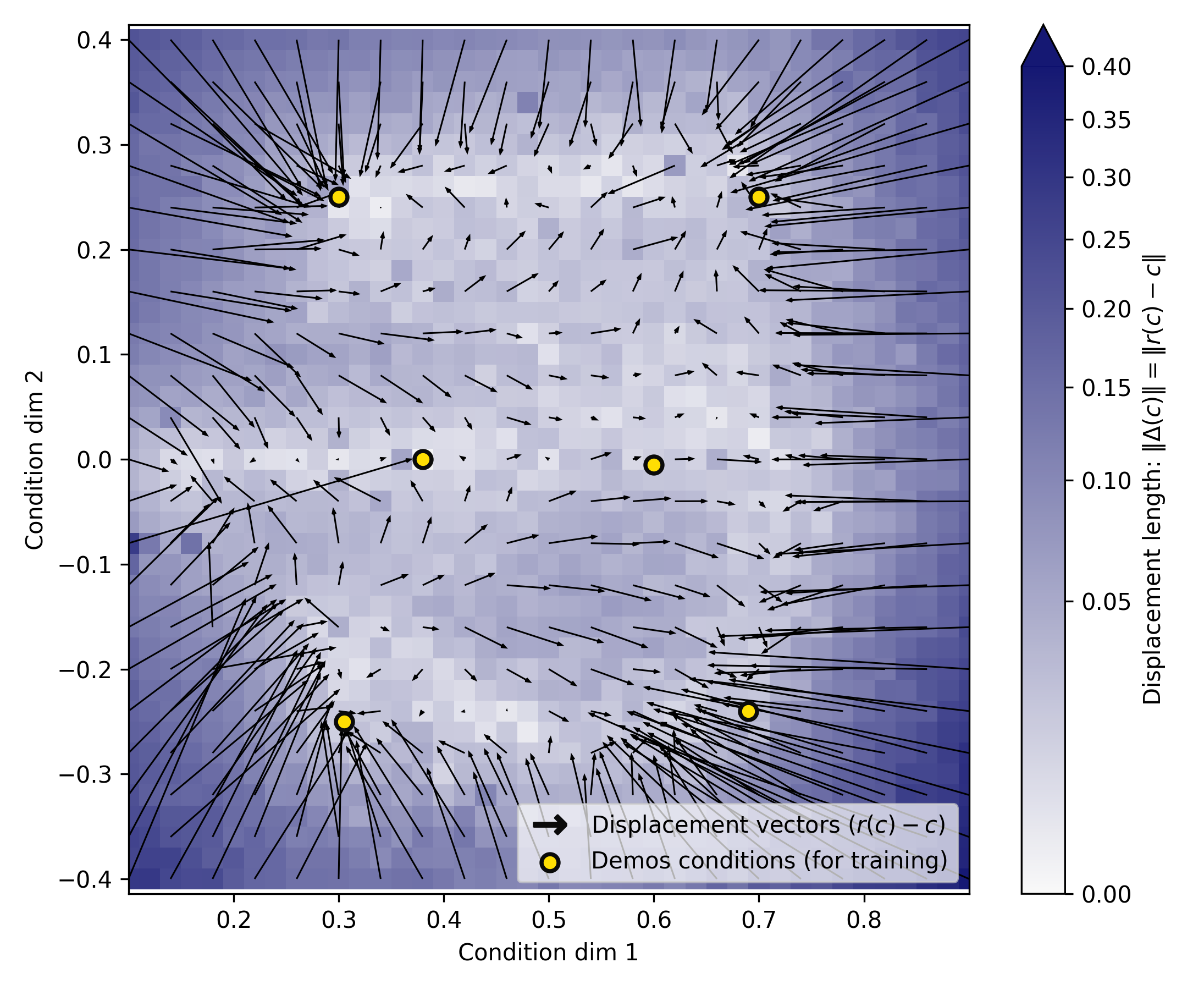}
        \vspace{-20pt}
        \caption{Full demonstration coverage (6 demos).}
        \label{fig:disp_full}
    \end{subfigure}

    \vspace{0.6em}

    \begin{subfigure}{0.7\linewidth}
        \centering
        \includegraphics[width=\linewidth]{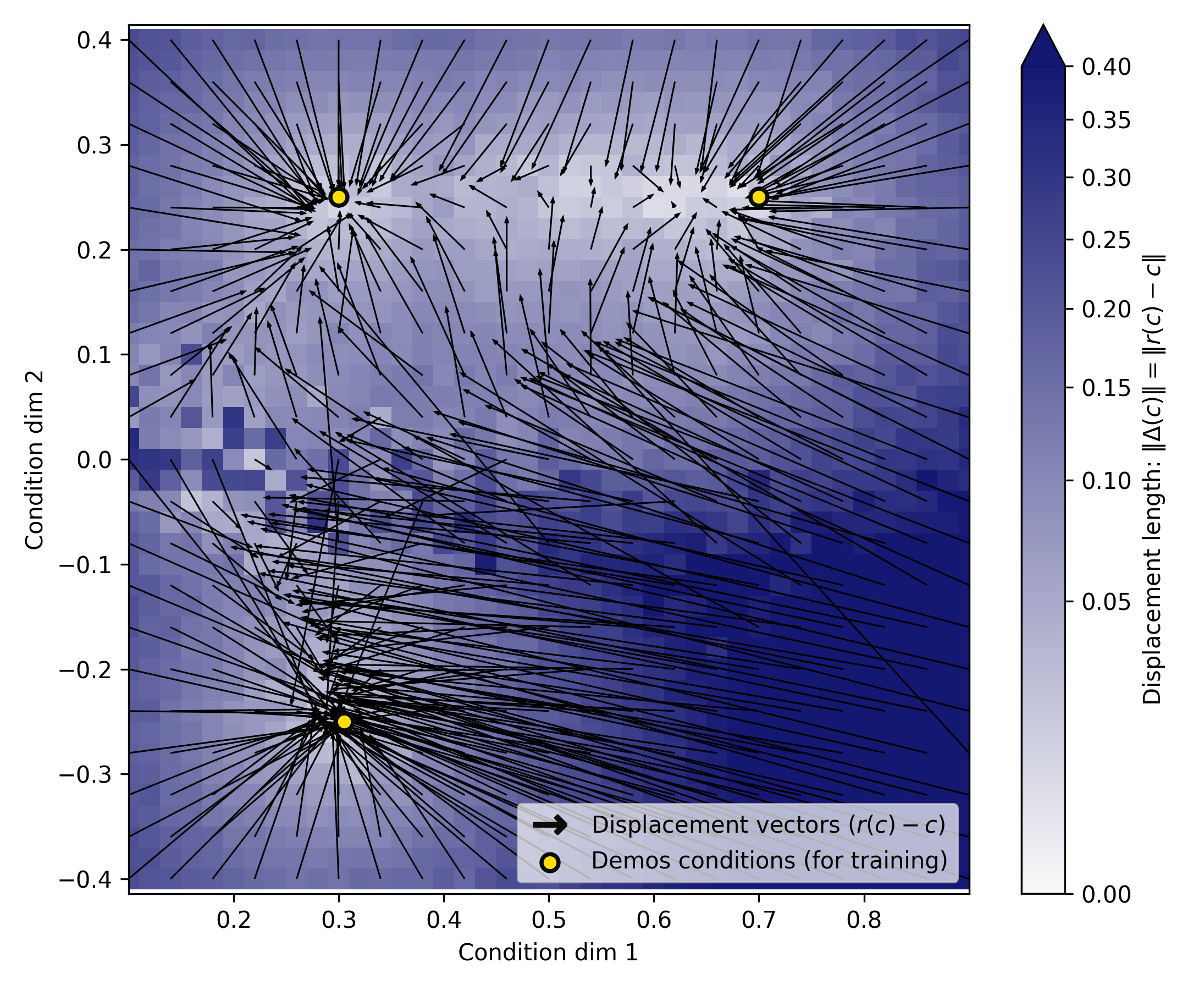}
        \vspace{-20pt}
        \caption{Reduced demonstration coverage (3 demos).}
        \label{fig:disp_reduced}
    \end{subfigure}

    \caption{Displacement field of \textit{Basketball Shot} in simulator (Equation~\ref{equ:delta}). Arrows indicate the $\Delta(\mathbf{c})$ vector and the heatmap indicates its scalar distance. Yellow markers are demonstration landing positions used for training. (a) With 6 demos, the mapping remains locally smooth within the demonstration-supported region. (b) With only 3 demos, high-error regions expand and local distortions increase, indicating degraded mapping stability.}
    \label{fig:disp_field}
    \vspace{-20pt}
\end{figure}

We provide additional experiments on the same \textit{Basketball Shot} in simulator, to visualize the correlation between action generation condition and the corresponding outcome.
Given the same trained generator $\pi$, we densely sample a wide region of the 2D condition space, and for each sampled condition $\mathbf{c}$, we generate a motion $\pi(\mathbf{c})$ and roll it out to get $\mathbf{r}(\mathbf{c})$. We visualize the outcome displacement vector
\begin{equation}
    \Delta(\mathbf{c}) =  \mathbf{r}(\mathbf{c}) - \mathbf{c} 
\label{equ:delta}
\end{equation}
and its magnitude $ \|\Delta(\mathbf{c})\|$ in Figure~\ref{fig:disp_field}, which captures the global condition-to-outcome mapping.

\textbf{Full demonstration coverage (6 demos).}
Figure~\ref{fig:disp_field}\subref{fig:disp_full} shows the displacement field when the diffusion prior is trained using all six demonstrations (yellow markers), where each marker corresponds to both the demonstrated landing position and the associated training condition. Within the demonstration-supported region, the displacement magnitude remains relatively small and the field exhibits locally smooth structure, while errors increase toward the boundary, indicating reduced reliability outside the covered range.

\textbf{Reduced demonstration coverage (3 demos).}
When the prior is trained with only three demonstrations (Figure~\ref{fig:disp_field}\subref{fig:disp_reduced}), high-error regions expand and the field becomes less structured, suggesting that insufficient demonstration spread can lead to unstable condition-to-outcome geometry.

Overall, the results indicate that effective adaptation does not require dense demonstration coverage, but does rely on a modest set of appropriately distributed priors to induce a smooth and structured condition-to-outcome geometry.

\section{Conclusion}
\label{sec:conclusion}

We introduced Prior Reinforce, a framework for low-trial goal-conditioned dynamic manipulation that separates motion generation from semantic outcome-driven adaptation. By operating in a learned low-dimensional condition space, the method enables efficient refinement under sparse and noisy feedback. Experiments across multiple real-world tasks and controlled simulations demonstrate that new goals can be reached within as few as ten total trials, while remaining robust to perception noise and moderate hardware variations. We hope this perspective — leveraging structured motion priors with low-dimensional condition adaptation — provides a practical direction for low-trial real-world robot learning.
\section{Limitations}
\label{sec:limitations}

The effectiveness of \ours relies on the presence of a coherent motion manifold induced by a small set of demonstrations. When demonstration coverage is insufficient or unevenly distributed, the induced condition-to-outcome mapping may become less reliable outside the supported region, as illustrated in Fig.~\ref{fig:disp_field}. Moreover, the current framework does not provide formal convergence guarantees and may not directly extend to tasks requiring long-horizon closed-loop control or higher-dimensional success representations (i.e., higher than a 2D plane). These factors define the practical scope of applicability of the proposed approach.

\bibliographystyle{IEEEtran} 
\bibliography{IEEEabrv, IEEEfull}{}

\end{document}